\documentclass[11pt,a4paper,logo]{lumia}
\usepackage[sort&compress]{natbib}
\definecolor{abstrabg}{HTML}{F2D1C5}
\definecolor{abstradark}{HTML}{D97757}
\usepackage{xurl,float,algorithm,algorithmic,multirow,subcaption,tikz,nicefrac}
\usetikzlibrary{calc,positioning,shapes,arrows.meta,fit,backgrounds}
\graphicspath{{figures/}}
\renewcommand{\absfont}{\linespread{1.0}\fontsize{10}{12}\selectfont}
\DeclareRobustCommand{\skillDAG}{\ifmmode\textnormal{SkillDAG}\else\textsc{SkillDAG}\fi}
\DeclareRobustCommand{\gos}{\ifmmode\textnormal{GoS}\else\textsc{GoS}\fi}
\DeclareRobustCommand{\skillnet}{\ifmmode\textnormal{SkillNet}\else\textsc{SkillNet}\fi}
\setheadertext{SkillDAG}
\correspondingemail{\emailicon~\href{mailto:yu_xingrui@a-star.edu.sg}{yu\_xingrui@a-star.edu.sg}\quad $^*$ Equal contribution.\quad $^\dagger$ Corresponding author.}
\title{SkillDAG: Self-Evolving Typed Skill Graphs for LLM Skill Selection at Scale}
\author{
Tong Bai$^{1,*}$\quad Zhenglin Wan$^{2,*}$\quad Pengfei Zhou$^{2,5}$\\
\bfseries Xingrui Yu$^{3,\dagger}$\quad Yang You$^{2}$\quad Ivor W.~Tsang$^{3,4}$\\
{\small $^{1}$College of Computer Science and Artificial Intelligence, Fudan University, China}\\
{\small $^{2}$Department of Computer Science, National University of Singapore, Singapore}\\
{\small $^{3}$CFAR and IAIC, Agency for Science, Technology and Research (A*STAR), Singapore}\\
{\small $^{4}$College of Computing and Data Science, Nanyang Technological University, Singapore}\\
{\small $^{5}$InfRec, Cardinal AI Lab}
}
\teaserfigure{
\vspace{-6pt}
\noindent\includegraphics[width=\linewidth]{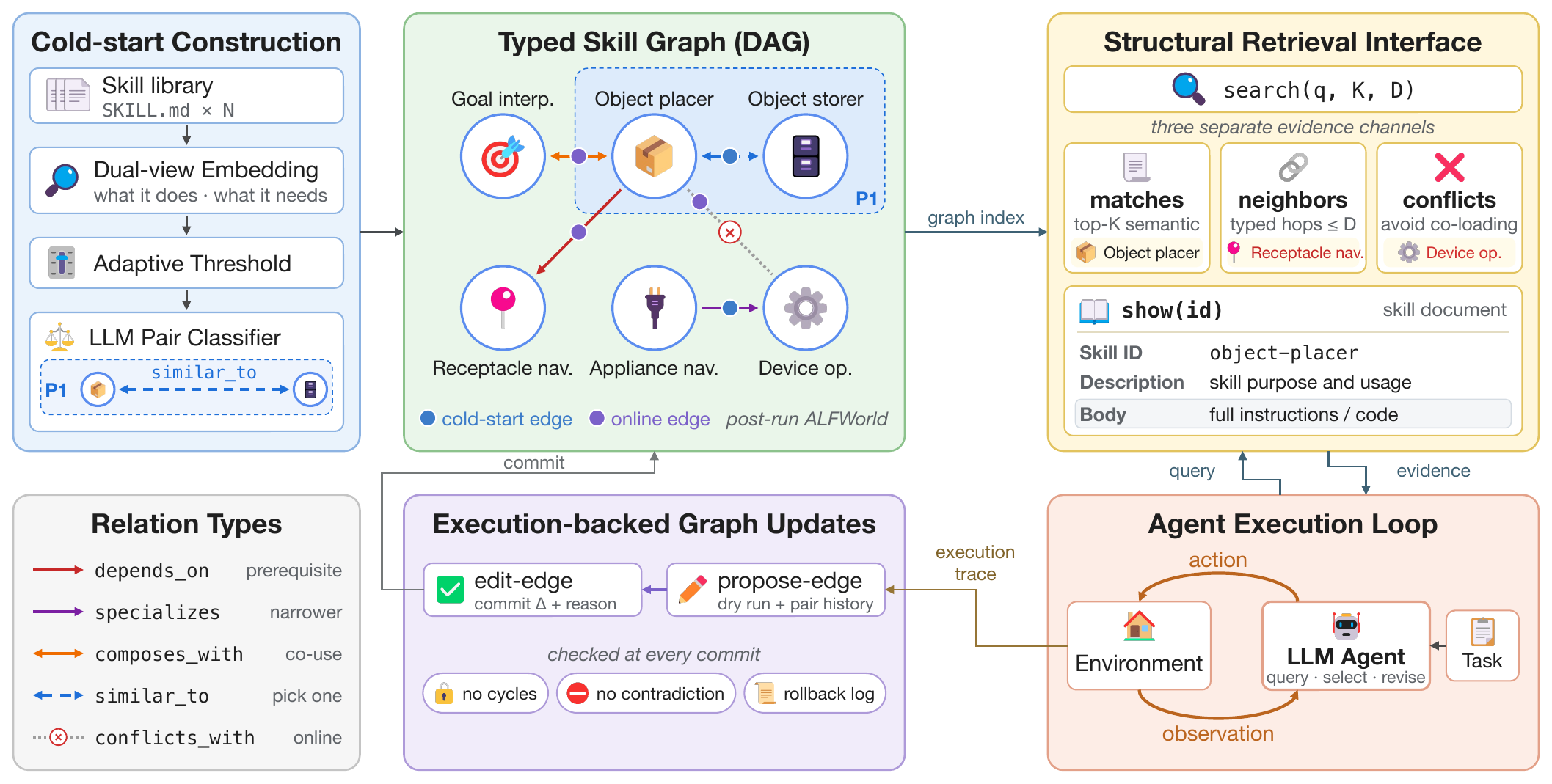}\par
\captionsetup{font=small,skip=3pt}
\captionof{figure}{\skillDAG{} overview. A two-view cold-start constructor initializes a typed skill graph. The agent queries semantic matches, typed neighbors, and conflicts, then loads skills and acts in the environment. Execution feedback supports proposed and validated graph updates, with an append-only history for rollback.}
\label{fig:skilldag-overview}
\vspace{-6pt}
}
\begin{document}
\begin{abstract}
Large skill libraries are now standard equipment for LLM agents, and which subset reaches the prompt decides what the agent can do.
Enumeration and top-$K$ embedding retrieval both treat skills as independent documents, assuming relevance is a property of one skill at a time. It is not: skills depend on, conflict with, specialize, and duplicate one another, so the closest match to a task can name it while omitting a prerequisite it needs. Graph retrieval encodes these relations but spends them in a fixed pipeline, handing the agent an opaque bundle.
We present \skillDAG{}, which exposes a typed skill graph to the LLM as an agent-callable interface rather than consuming it for the agent: \texttt{search} returns vector matches, typed-edge neighbors, and conflict signals, and a propose-then-commit protocol lets it register execution-backed edges as it works.
Across two benchmarks and three model families, \skillDAG{} outperforms full-library prompting, vector retrieval, and Graph-of-Skills, gaining $+14.3$ points of ALFWorld success and tying only at ceiling; its ranking holds flat across a $10\times$ pool expansion that degrades the fixed pipeline.
Inter-skill structure, these results suggest, is worth more to an agent that can query and revise it than to a retriever that consumes it.

\noindent\textbf{Code and data:} \begingroup\def\UrlBreaks{\do\/}\url{https://github.com/Ericbai06/SkillDAG}\endgroup

\end{abstract}
\maketitle
\section{Introduction}
\label{sec:introduction}

Large language model agents increasingly rely on external skill libraries to extend their capabilities beyond the procedural knowledge in model weights.
A skill is a self-contained package: a unique identifier, a natural-language description of when it applies, and a body of instructions or code the agent runs when invoked, often with auxiliary scripts and prerequisite notes~\citep{li2026gos}.
Voyager~\citep{wang2023voyager}, JARVIS-1~\citep{wang2023jarvis}, and ToolLLM~\citep{qin2024toolbench} report large gains once an agent is equipped with a curated skill collection.
As collections grow to thousands of entries, the bottleneck shifts from authoring to skill \emph{selection}: which subset to load cannot be decided from a name or embedding alone, and retrieval over large tool collections is itself a measurable failure mode of LLM agents~\citep{shi2025toolret}.

Two strategies dominate practice.
Concatenating the whole library is feasible only at small scale: token cost grows linearly and descriptions are lost in an overloaded prompt~\citep{liu-etal-2024-lost}.
Flat retrieval instead embeds the task and concatenates the top-$K$ skills by cosine similarity~\citep{karpukhin2020dpr,reimers2019sentencebert} or BM25~\citep{robertson2009bm25}.
Both hide the library's relational structure: the top match names the task while a functionally necessary prerequisite (a parser, converter, or setup utility) is silently omitted; analogous gaps arise from skills that interfere when co-selected, near-duplicates that waste context budget, and pairs that jointly enable a capability neither provides alone.
Curation can mask these failures at small scale, but its cost grows with the library.

Prior work models inter-skill structure but stops short of letting the LLM reason over it directly.
\skillnet{}~\citep{skillnet2025} organizes over 500{,}000 curated skills as a graph, but its graph relations are not updated online from execution evidence.
Graph of Skills~\citep[\gos{};][]{li2026gos} runs Personalized PageRank~\citep{page1999pagerank,haveliwala2003topicsensitive} over a typed skill graph at inference time and reports gains on SkillsBench and ALFWorld, but computes the bundle through a fixed seeding--diffusion--reranking pipeline and hands the agent opaque context: it cannot ask why a skill was chosen, retract a noisy retrieval, or record a relationship the offline pipeline missed.
XSkill~\citep{xskill2025} revises individual skills and CUA-Skill~\citep{cuaskill2025} composes them over an offline-authored graph; neither learns the relations themselves from execution.

We propose \skillDAG{}, built on the inverse design choice: rather than the substrate of a fixed retrieval algorithm, the graph is exposed to the LLM as an inference-time, agent-callable structural retrieval interface, on the same footing as the tools the agent already uses for execution.
The LLM, not a graph-ranking policy, decides how structural evidence should affect execution; the graph in turn accumulates structural knowledge across episodes through evidence-backed agent edits.

\noindent Our contributions are summarized as follows:
\begin{itemize}
\setlength{\itemsep}{2pt}\setlength{\parsep}{0pt}\setlength{\topsep}{2pt}
\item We identify that flat retrieval over large skill libraries silently omits structurally necessary skills and concatenates redundant or interfering ones---a failure mode invisible to similarity matching alone.
\item We propose \skillDAG{}, an agent-callable structural retrieval interface that exposes a typed skill relation graph to the LLM, replacing a fixed graph-ranking policy with structural evidence the agent reasons over directly.
\item We design an online propose-then-commit edit protocol under three structural invariants (acyclicity, non-contradiction, append-only reversibility) that grows the graph from single-episode evidence, paired with a two-view cold-start constructor that recovers cross-functional pairs flat self-similarity misses.
\item Experiments on ALFWorld~\citep{shridhar2021alfworld} and SkillsBench~\citep{li2026gos} with MiniMax-M2.7 and \texttt{gpt-5.2-codex} show \skillDAG{} reaches 67.1\%/93.6\% ALFWorld success and 27.3\%/36.8\% SkillsBench reward, with Ret@5 of 78.2 stable across a $10\times$ skill-pool expansion.
\end{itemize}

\section{Related Work}
\label{sec:related}

\paragraph{Skill and tool libraries.}
LLM agents are augmented with external capabilities exposed as tools~\citep{schick2023toolformer,qin2024toolbench,patil2024gorilla}, skills~\citep{wang2023voyager,wang2023jarvis}, code actions~\citep{liang2023codeaspolicies,wang2024codeact}, or grounded primitives~\citep{ahn2022saycan,huang2022innermonologue}.
As libraries scale, selection rather than authoring becomes the bottleneck: ToolkenGPT~\citep{hao2023toolkengpt} learns per-tool embeddings, AnyTool~\citep{du2024anytool} adds a hierarchical API retriever, CRAFT~\citep{yuan2024craft} creates and retrieves specialized tool-sets, and library-learning methods compress trajectories into documented reusable code~\citep{grand2024lilo}.
\skillnet{}~\citep{skillnet2025} organizes over 500K curated skills as a graph consumed by its Wiki-based routing layer, CUA-Skill~\citep{cuaskill2025} pairs skills with parameterized execution and composition graphs, and GraSP~\citep{chen2026grasp} compiles retrieved skills into typed DAGs and repairs them during execution; in each case the structure is consumed by a fixed pipeline rather than exposed to the agent as a queryable, editable interface, so the agent itself cannot resolve prerequisites or conflicts as the library grows.

\paragraph{Skill retrieval and structured RAG.}
Beyond flat top-$K$ RAG~\citep{lewis2020rag}, recent work injects structure: GraphRAG~\citep{edge2024graphrag} indexes a community graph for offline summarization, HuggingGPT~\citep{shen2023hugginggpt} composes models by task decomposition over a flat catalog, HippoRAG~\citep{gutierrez2024hipporag} retrieves by Personalized PageRank over a KG memory, LightRAG~\citep{guo2025lightrag} maintains an incrementally updated graph index, and GeAR~\citep{shen2025gear} performs agent-driven graph expansion.
For tools specifically, Re-Invoke~\citep{chen2024reinvoke} rewrites queries for zero-shot retrieval and ToolGen~\citep{wang2025toolgen} folds lookup into generation via tool tokens.
Graph of Skills~\citep[\gos{};][]{li2026gos}, the closest baseline, builds a typed skill graph offline and retrieves a bundle by reverse-aware Personalized PageRank that is concatenated into context as opaque text.
\skillDAG{} inverts two choices: the graph is an agent-callable structural retrieval interface rather than a fixed retriever's substrate, and its edges are editable during execution rather than frozen at construction.

\paragraph{Self-evolving agent.}
ReAct~\citep{yao2023react} interleaves reasoning~\citep{wei2022cot} and action, extended by deliberate search~\citep{yao2023tot}; Reflexion~\citep{shinn2023reflexion} and Self-Refine~\citep{madaan2023selfrefine} add verbal self-correction; ExpeL~\citep{zhao2024expel}, AutoAct~\citep{qiao2024autoact}, and Agent-Pro~\citep{zhang2024agentpro} distill experience into reusable insights or policies.
Closer to skill accumulation, Agent Workflow Memory~\citep{wang2024awm} induces reusable workflows online, ICAL~\citep{sarch2024ical} and STE~\citep{wang2024ste} abstract trajectories into memory, AutoGuide~\citep{fu2024autoguide} retrieves state-conditioned guidelines, and OS-Copilot~\citep{wu2024oscopilot} accumulates skills in a tool-rich OS; surveys frame these as non-parametric self-evolution~\citep{gao2025selfevolving}.
Generative Agents~\citep{park2023generative} and MemGPT~\citep{packer2023memgpt} manage free-text memory, and XSkill~\citep{xskill2025} updates individual skills.
Our propose/edit-edge mechanism instead mutates a typed relational structure shared by all future agents, enabling precise, auditable updates.

\paragraph{Agent benchmarks.}
We evaluate on ALFWorld~\citep{shridhar2021alfworld} and SkillsBench~\citep{li2026gos}, within a landscape spanning cross-domain tasks~\citep{liu2024agentbench}, web environments~\citep{zhou2024webarena}, general assistants~\citep{mialon2023gaia}, desktop/OS control~\citep{xie2024osworld}, API-rich apps~\citep{trivedi2024appworld}, tool-agent-user dialogue~\citep{yao2024taubench}, and function calling~\citep{patil2025bfcl}.
These probe agent capability broadly rather than skill selection specifically; the pressure motivating \skillDAG{} appears wherever the callable surface itself grows large, which \citet{shi2025toolret} documents as a measurable retrieval failure mode.

\section{Method}
\label{sec:method}

\subsection{Overview}
\label{sec:method-overview}

\skillDAG{} starts from a simple inversion: the typed graph is not hidden inside a fixed retrieval pipeline, but exposed as an agent-callable interface the LLM can query, inspect, and revise during execution. The typed relation graph gives otherwise flat skill libraries operational structure: prerequisite and specialization edges organize reusable progressions, composition and similarity edges surface adjacent skills, and conflict edges mark combinations the agent should avoid. The \texttt{search}/\texttt{show} interface turns that structure into action by returning semantic matches, typed neighbors, and conflicts as separate evidence channels while loading full skill bodies only on demand. To initialize such structure before any episode runs, the cold-start constructor links skills from two complementary views: what a skill does and what it needs. Static structure is inevitably incomplete, however, so the online protocol lets the agent propose and commit execution-backed edits while acyclicity, non-contradiction, and append-only reversibility keep the graph from drifting into incoherence.

\subsection{Skill Relation Graph Definition}
\label{sec:graph-def}

Prerequisites, alternatives, and conflicts collapse onto a single cosine score under the default retrieval baseline (rank skills by self-description similarity and concatenate the top-$K$), so a hard prerequisite and an interchangeable alternative both surface as ``high-cosine neighbors,'' even though loading the former unlocks the task while loading the latter wastes context and dilutes the agent's attention. To prevent this collapse, we elevate the relation \emph{type} itself to be part of the agent-facing semantics, so the LLM acts on \emph{what kind of} relation it sees rather than merely \emph{how similar} two skills look. We instantiate this principle as a typed graph. Let the skill library be a finite set $\mathcal{C} = \{s_1, \ldots, s_n\}$ of skill packages, where $s_i$ is also the unique identifier; the graph $G = (V, E, \tau)$ has $V = \mathcal{C}$ and a typing function $\tau$ assigning each edge one of five operationally specified types: \texttt{depends\_on}$(A,B)$ marks $A$ as requiring the prerequisite $B$; \texttt{specializes}$(D,A)$ marks $D$ as the narrower variant to prefer over $A$; \texttt{composes\_with} marks synergistic co-use; \texttt{similar\_to} marks functional redundancy (pick one); and \texttt{conflicts\_with} marks pairs that should not be co-selected.

\begin{figure*}[!t]
\centering
\includegraphics[width=\linewidth]{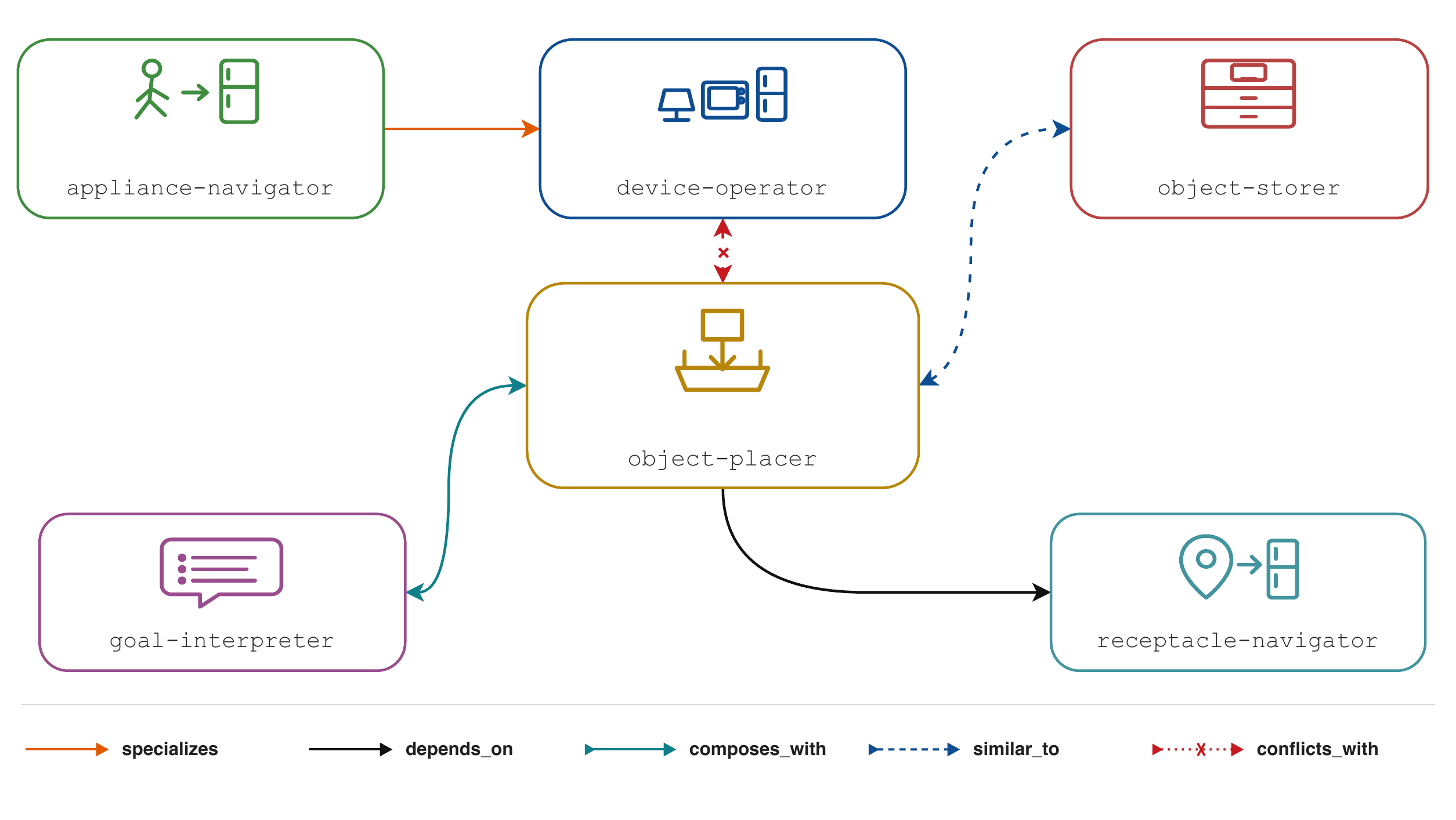}
\caption{A six-node induced subgraph from a post-run ALFWorld graph containing one edge of each type: cold-start \texttt{specializes}/\texttt{similar\_to} and online \texttt{depends\_on}/\texttt{composes\_with}/\texttt{conflicts\_with}.}
\label{fig:typed-edge-graph}
\end{figure*}

Two operational constraints make this typed graph usable as an agent-facing interface. \emph{Conflicts as a pruning signal, not a navigation edge.} We deliberately exclude \texttt{conflicts\_with} from traversal because navigating a conflict edge would actively undermine retrieval by surfacing the very skills the LLM has been told to avoid; the four positive types are walked by the retrieval interface, while \texttt{conflicts\_with} is only used to exclude co-selection (\S\ref{sec:graph-interface}). \emph{Acyclicity on the backbone.} We reserve \texttt{depends\_on} and \texttt{specializes} as the directed backbone whose acyclicity gives \skillDAG{} its name; the invariant is enforced at commit time (\S\ref{sec:online-edit}). Figure~\ref{fig:typed-edge-graph} shows a six-node post-run ALFWorld subgraph exercising all five types.

\subsection{Agent-Callable Structural Retrieval Interface}
\label{sec:graph-interface}

The default move in prior graph retrievers is to collapse typed structure back into one ranked bundle, but doing so would undo the work of \S\ref{sec:graph-def}, hiding from the LLM exactly the distinctions we built the graph to preserve. We therefore designed \texttt{search} as an agent-callable interface that returns semantic matches, typed neighbors, and conflicts as three separate evidence channels the LLM reasons over independently.

A single \texttt{search}$(q, K, D)$ call returns three fields, each answering a distinct question the LLM would otherwise have to ask separately: \textbf{matches} (``what is topically close to $q$?''), the top-$K$ skills by query--node embedding cosine~\citep{karpukhin2020dpr}; \textbf{neighbors} (``what does the typed graph attach to those matches?''), skills reached by bounded breadth-first traversal of depth $D$ over the four walkable edge types; and \textbf{conflicts} (``what should I not co-load with them?''), the one-hop \texttt{conflicts\_with} edges incident to any match.

We chose to keep these channels separate, rather than fusing them into a single score, because the decomposition gives the agent three capabilities that a ranked bundle cannot offer: it can drop an entire field under a tight context budget (e.g., skip \textbf{neighbors} when context is scarce), schedule \texttt{search} calls on its own initiative (e.g., re-query after partial progress), and leave a reasoning trace documenting why each skill was selected or excluded.

\subsection{Cold-Start Graph Construction}
\label{sec:cold-start}

The interface of \S\ref{sec:graph-interface} assumes a graph already exists. The obvious choice (embed each skill's self-description and link nearest neighbors) has a specific failure mode we set out to fix: skills that should be connected often describe different actions, where one provides what another requires. A cooling skill and a pickup skill, for example, have disjoint self-descriptions, yet the cooling skill requires ``object in hand'' produced by the pickup skill. Self-description similarity alone systematically misses such cross-functional pairs, so we designed a two-view embedding strategy that initializes the graph from static \texttt{SKILL.md} text while explicitly recovering these bridges.

The key departure from single-view retrieval is the second embedding. For each skill $v$, $e_{\text{self}}(v)$ captures \emph{what a skill does} (identifier, description, body preview), while $e_{\text{needs}}(v)$ captures \emph{what it requires}: an LLM imagines a few invoking tasks and summarizes the shared prerequisites in one sentence (HyDE-style~\citep{gao2023hyde}), then embeds the summary. $e_{\text{self}}$ surfaces topical neighbors; $e_{\text{needs}}$ surfaces functional bridges, and a candidate pair appears if either ranking exceeds the adaptive cosine threshold. An LLM classifier~\citep{zheng2023llmasjudge} then assigns one of the four static types or \texttt{none}. Figure~\ref{fig:skilldag-overview} places this cold-start construction within the full retrieval and graph-evolution loop.

We reserve \texttt{conflicts\_with} for the online protocol (\S\ref{sec:online-edit}) by operational necessity: a conflict is, by definition, a pair whose co-use predictably degrades success, and that evidence simply does not exist until execution is observed. No amount of static reading recovers it.

\subsection{Online Graph Evolution}
\label{sec:online-edit}

Even a well-constructed cold-start graph is necessarily incomplete. By the argument just made it cannot contain \texttt{conflicts\_with} edges at all; more broadly, many useful relations appear only during execution: a prerequisite the cold-start LLM did not anticipate, an interference pair observable only when both skills are loaded together, a refinement of an existing relation after a near-miss. To close this gap, we make graph construction an online part of agency rather than offline preprocessing: the per-episode loop includes a \texttt{propose-edge}/\texttt{edit-edge} pair through which the agent commits evidence-backed edits at the same pace it acts in the environment.

We split the edit into two calls so that committing is always a deliberate act rather than a side effect of execution. \texttt{propose-edge} performs a dry run: it previews the change and surfaces existing edges and history for the same skill pair, so the agent can reason about whether to commit. \texttt{edit-edge} then commits the change with a natural-language \texttt{reason} and the supporting execution trace; the same call can add, delete, or retype an existing edge, giving the agent an explicit retraction path for stale or mistaken relations. We deliberately chose to gate commits only by the agent's own reasoning, not by a downstream voting threshold or a human curator, so that a single execution episode suffices and the graph evolves at agent speed rather than curation speed. The mixed-origin subgraph in Figure~\ref{fig:typed-edge-graph} illustrates the state this preview must expose: online relations accumulate alongside cold-start structure, so each proposed change must be evaluated against the existing edges and history for its pair.

Gating commits only by the agent's reasoning raises the risk of single-shot corruption, which we bound with three structural invariants every commit must pass; the invariants are chosen to limit the blast radius of a bad edit without second-guessing the agent's semantic judgment. \emph{Acyclicity}: a \texttt{depends\_on}/\texttt{specializes} commit that would close a backbone cycle is rejected, since a circular prerequisite chain would loop the agent through skills that each demand the next, voiding the DAG property declared in \S\ref{sec:graph-def}. \emph{Non-contradiction}: a pair cannot simultaneously carry \texttt{conflicts\_with} and a positive surfacing edge, preventing the navigation and pruning signals from cancelling each other. \emph{Reversibility}: an append-only history log supports rollback by recency, so a spurious edit is bounded and recoverable. These invariants are structural rather than semantic by design: they preserve the LLM's authority over what each edge \emph{means} while preventing the graph from drifting into incoherence.

\section{Experiments}
\label{sec:experiments}
Our evaluation asks when and why the interface design improves end-task performance. We first compare \skillDAG{} against full-library prompting, vector retrieval, and \gos{} on task reward, then replay identical queries through both retrievers to isolate pre-execution retrieval quality. Controlled Flat/Frozen/\skillDAG{} runs separate the interface, typed graph, and online-edit conditions. MiniMax-M2.7, \texttt{gpt-5.2-codex}, and \texttt{deepseek-v4-flash} test backbone sensitivity, while a $10\times$ skill-pool expansion, randomized task orders, and an ALFWorld train/test split test scale and transfer.

\begin{table*}[!t]
\centering
\footnotesize
\setlength{\tabcolsep}{4pt}
\begin{tabular}{@{}ll cccc cccc@{}}
\toprule
& & \multicolumn{4}{c}{\textbf{SkillsBench}} & \multicolumn{4}{c}{\textbf{ALFWorld}} \\
\cmidrule(lr){3-6} \cmidrule(lr){7-10}
\textbf{Model} & \textbf{Method} & R$\uparrow$ & Ret@5$\uparrow$ & Ret@1$\uparrow$ & MRR$\uparrow$ & R$\uparrow$ & Ret@5$\uparrow$ & Ret@1$\uparrow$ & MRR$\uparrow$ \\
\midrule
\multirow{4}{*}{MiniMax-M2.7}
 & Vanilla Skills & 17.2$^\dagger$ & --            & --            & --            & 47.1$^\dagger$          & --            & --            & --            \\
 & Vector Skills  & 10.4$^\dagger$ & 10.8          & 3.6           & 5.8           & 50.7$^\dagger$          & 68.6          & 37.9          & 49.2          \\
 & \gos{}         & 18.7$^\dagger$ & 65.5          & 50.6          & 57.3          & 52.9                    & 86.4          & 56.4          & 67.9          \\
 & \skillDAG{}    & \textbf{27.3}  & \textbf{78.2} & \textbf{66.7} & \textbf{71.3} & \textbf{67.1}           & \textbf{92.1} & \textbf{57.9} & \textbf{71.1} \\
\midrule
\multirow{4}{*}{gpt-5.2-codex}
 & Vanilla Skills & 27.4$^\dagger$ & --            & --            & --            & 89.3$^\dagger$          & --            & --            & --            \\
 & Vector Skills  & 21.5$^\dagger$ & 10.8          & 3.6           & 5.8           & 92.9$^\dagger$          & 68.6          & 37.9          & 49.2          \\
 & \gos{}         & 34.4$^\dagger$ & 65.5          & 50.6          & 57.3          & \textbf{93.6}$^\dagger$ & \textbf{86.4} & 56.4          & 67.9          \\
 & \skillDAG{}    & \textbf{36.8}  & \textbf{75.9} & \textbf{70.1} & \textbf{73.0} & \textbf{93.6}           & 85.1          & \textbf{60.4} & \textbf{69.6} \\
\bottomrule
\end{tabular}
\caption{Main results. \texttt{R} = reward (\%); ALFWorld \texttt{R} = success rate. SkillsBench \texttt{R} at 1000-skill scale. \texttt{Ret@5} uses $K=5$ throughout; \texttt{Ret@5}/\texttt{Ret@1}/MRR ($\uparrow$) are intrinsic retrieval metrics computed by us. $\dagger$~from \citet{li2026gos} Table~1; every unmarked value is our own run. Best per model block in bold.}
\label{tab:main-results}
\end{table*}

\subsection{Experimental Setup}

\noindent\textbf{Benchmarks.}
We evaluate on two benchmarks used by \citet{li2026gos}, which exercise different aspects of skill selection at scale.
\emph{ALFWorld}~\citep{shridhar2021alfworld} is a text-based embodied environment whose six task types (pick-and-place, examine-in-light, clean-, heat-, cool-and-place, pick-two-and-place) require composing navigation, manipulation, and state-change behaviors; we run the 140-episode \texttt{valid\_seen} split with a 37-skill procedural library.
\emph{SkillsBench}~\citep{li2026gos} is 87 containerized code-generation tasks scored by unit-test pass fraction, shipped at four skill-pool scales (200/500/1000/2000 \texttt{SKILL.md} documents) so retrieval quality can be studied as the pool grows.

\noindent\textbf{Baselines.}
We compare against three baselines from \citet{li2026gos}.
\emph{Vanilla Skills} places the entire library in the prompt and acts via ReAct~\citep{yao2023react} (the recall upper bound and context-cost reference).
\emph{Vector Skills} concatenates the top-$K$ embedding-cosine matches, using no graph structure.
\emph{Graph of Skills} (\gos{})~\citep{li2026gos} retrieves a bundle by hybrid semantic--lexical seeding plus reverse-aware Personalized PageRank over a typed graph, reranked under a token budget, then concatenated into the prompt.
\skillDAG{} (ours) exposes the graph as an agent-callable structural retrieval interface: the agent issues \texttt{search}/\texttt{show}/\texttt{propose-edge}/\texttt{edit-edge} during execution, acts on the three-field search response, and commits evidence-backed edges under the guardrail checks.

\noindent\textbf{Models \& evaluation.}
We report per-episode reward \texttt{R} (\%); for ALFWorld \texttt{R} equals success rate.
The intrinsic retrieval columns are computed by us on identical inputs: the actual \texttt{search} queries issued by the \skillDAG{} agent during its 87-task run, evaluated by each pipeline on the scale-1000 pool ($n{=}140$ ALFWorld episodes, $n{=}87$ SkillsBench tasks).
We additionally ran an extended four-condition experiment on \texttt{deepseek-v4-flash} over the same 140 ALFWorld episodes (Figure~\ref{fig:component-ablation}b).

\subsection{Main Results}
\label{sec:results}
Table~\ref{tab:main-results} shows \skillDAG{} improves most under MiniMax-M2.7: ALFWorld $52.9\!\to\!67.1$ (74/140 versus 94/140, $+14.3$) and SkillsBench $18.7\!\to\!27.3$ ($+8.6$). Intrinsic SkillsBench retrieval likewise improves over \gos{} on the same MiniMax query distribution: Ret@5 $65.5\!\to\!78.2$, Ret@1 $50.6\!\to\!66.7$, MRR $57.3\!\to\!71.3$, isolating a retrieval-pipeline gain before agent-side execution; with embedding and queries held fixed, \skillDAG{}'s \texttt{matches}-field cosine over typed-graph node embeddings ranks more accurately than \gos{}'s HNSW~\citep{malkov2020hnsw} seed${+}$PPR~\citep{page1999pagerank} diffusion.

On \texttt{gpt-5.2-codex}, \skillDAG{} ties \gos{} on ALFWorld ($93.6\%$), where both methods are close to the task ceiling, but remains best on SkillsBench ($34.4\!\to\!36.8\%$) and yields $75.9/70.1/73.0$ Ret@5/Ret@1/MRR on codex trajectory queries. The \texttt{deepseek-v4-flash} batch in Figure~\ref{fig:component-ablation}b shows a larger \skillDAG{}-over-\gos{} ALFWorld gain ($75.0\!\to\!82.9$). Taken together, \skillDAG{} leads on both benchmarks under MiniMax-M2.7 and on SkillsBench under \texttt{gpt-5.2-codex}, where ALFWorld is already at ceiling, while its ranked-matches retrieval stays ahead of \gos{} at every pool scale across a $10\times$ expansion.

\subsection{Scale and Edit-Transfer Ablations}
\label{sec:ablations}
We isolate two effects behind the main results: whether the agent's online edits persist as durable, recall-safe structure, and whether the typed-graph retriever stays robust as the skill pool grows.

\begin{figure}[!htbp]
\centering
\begin{subfigure}[t]{0.36\textwidth}
  \centering
  \includegraphics[width=\linewidth]{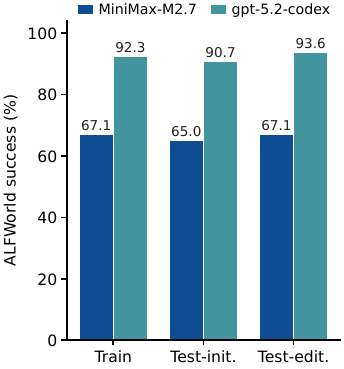}
  \caption{ALFWorld train/test.}
  \label{fig:alfworld-traintest}
\end{subfigure}
\hfill
\begin{subfigure}[t]{0.62\textwidth}
  \centering
  \includegraphics[width=\linewidth]{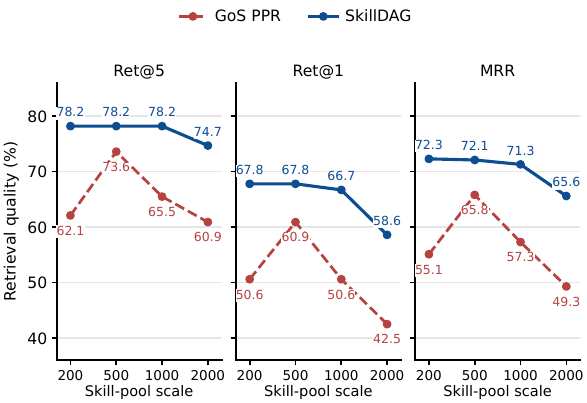}
  \caption{Cross-scale retrieval on SkillsBench.}
  \label{fig:cross-scale-retrieval}
\end{subfigure}
\caption{Robustness experiments. (a) ALFWorld success (\%) across Train (420 in-domain, edits on), Test-initial (140 held-out on cold-start graph), and Test-edited (same 140 on train-produced graph), for both backbones. (b) Same 87 \skillDAG{} queries through both retrievers, shared \texttt{text-embedding-3-large}; \skillDAG{} outranks \gos{} at every scale and stays flat across the $10\times$ expansion; scale-1000 matches the MiniMax row of Table~\ref{tab:main-results}.}
\label{fig:robustness}
\end{figure}

We measure online editing two ways: an ALFWorld train/test split that tests whether committed edges transfer, and a SkillsBench cold-vs-edited query replay that tests what an accepted edge does to retrieved recall (Table~\ref{tab:edit-effect}).

\begin{table}[tb]
\centering
\begin{tabular}{@{}lcc@{}}
\toprule
Graph & Edges & Mean GT/q \\
\midrule
Cold-start          & 2{,}359 & 1.915          \\
Edited (+27 online) & 2{,}386 & \textbf{1.984} \\
\bottomrule
\end{tabular}
\caption{SkillsBench scale-1000 retrieval, replaying 316 \texttt{search} queries from 65 tasks against the cold-start vs.\ edited graph. ``Mean GT/q'' is the average number of ground-truth skills retrieved per query (mean ground-truth size 2.52 skills/task).}
\label{tab:edit-effect}
\end{table}

\Needspace{8\baselineskip}
\noindent\textbf{Committed edges persist as durable structure.}
A 420-episode in-domain \emph{train} run accumulates edits; \emph{test-initial} (140 held-out episodes, cold-start graph) and \emph{test-edited} (same 140, train-produced graph) are then run with edits disabled.
On both backbones train and test-initial sit within $2.1$ points (MiniMax-M2.7 $67.1/65.0/67.1\%$; \texttt{gpt-5.2-codex} $92.3/90.7/93.6\%$; Figure~\ref{fig:alfworld-traintest}): transfer holds to the held-out split, and test-edited recovers the train number, so committed edges persist as durable structure rather than training-time stochastics.

\noindent\textbf{Online edits expand ground-truth recall set-monotonically.}
Replaying all 316 scale-1000 \texttt{search} queries, the edited graph (+27 single-observation commits) retrieves more ground-truth skills per query ($1.984$ vs.\ $1.915$; Table~\ref{tab:edit-effect}) and \emph{zero} queries lose any ground-truth retrieval.
This follows by construction: the scorer takes the untruncated union of \texttt{matches} and BFS neighbors, so an accepted edge can only enlarge the retrieved set, never evicting a ground-truth skill the cold-start graph already surfaced; this is a recall property, not a context-budget claim.

\noindent\textbf{\skillDAG{} ranking stays robust as the pool grows $10\times$.}
With the embedding backbone held fixed, \skillDAG{}'s plain \texttt{matches} cosine outranks \gos{} on every retrieval metric at every pool scale despite using no diffusion step: $+16.1/{+}17.2/{+}17.2$ (Ret@5/Ret@1/MRR) at 200 skills and $+13.8/{+}16.1/{+}16.3$ at 2000, narrowest at 500 ($+4.6/{+}6.9/{+}6.3$) where \gos{}'s PPR pull happens to align with the gold target; across the $10\times$ pool expansion it loses only $3.5$ Ret@5 points ($78.2\!\rightarrow\!74.7$), scaling robustly to larger libraries (Figure~\ref{fig:cross-scale-retrieval}).
This isolates the offline retrieval pipeline, not the typed-edge \texttt{neighbors}/\texttt{conflicts} fields the agent exercises during execution.

\subsection{Component Ablation and Backbone Generalization}
\label{sec:component-ablation}

\begin{figure}[!htbp]
\centering
\includegraphics[width=0.60\linewidth]{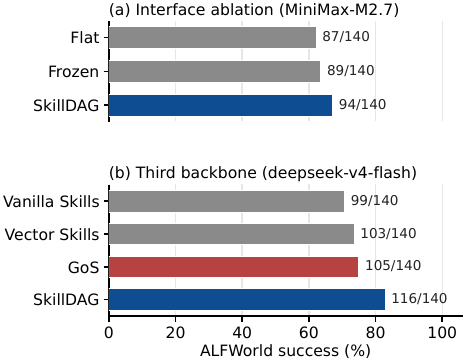}
\caption{ALFWorld success over 140 episodes. (a) Controlled MiniMax-M2.7 interface ablation: Flat exposes vector matches without typed edges (87/140), Frozen enables typed-graph retrieval without online edits (89/140), and \skillDAG{} adds post-episode reflection and editing (94/140). (b) Third-backbone \texttt{deepseek-v4-flash} batch, four conditions with identical episode provenance across methods.}
\label{fig:component-ablation}
\end{figure}

Figure~\ref{fig:component-ablation}a ablates the interface itself. All three conditions hold the task indices, 30-step cap, node cache, capability prefix, and evaluator fixed, so they differ only in what the agent may see and do: Flat exposes vector matches alone, Frozen adds typed-graph retrieval, and \skillDAG{} additionally commits post-episode edits. Flat reaches 87/140, Frozen reaches 89/140, and \skillDAG{} reaches 94/140. Frozen improves over Flat by 1.43 points, isolating the typed-graph contribution; \skillDAG{} improves over Frozen by 3.57 points, isolating post-episode editing; and the combined \skillDAG{}-over-Flat gain is 5.00 points. In the held-out transfer run summarized above, the MiniMax cold-start and train-edited graph states score 91/140 and 94/140, respectively, with further test-time editing disabled.

Task-order ablations provide a separate robustness check. \skillDAG{} scores 94/140 in natural order and 93/140 and 95/140 under random seeds. Across all 420 episodes, success is 282/420 (67.14\%) with a 1.43-point range. The result supports order robustness; it does not treat early/late position as learning progress because position remains confounded with task type.

Figure~\ref{fig:component-ablation}b extends the comparison to a third backbone family. \skillDAG{} reaches 116/140, versus 99/140 for Vanilla, 103/140 for Vector, and 105/140 for \gos{}, leading the three baselines by 12.14, 9.29, and 7.86 points. This batch covers ALFWorld only. Across backbones, the ALFWorld tie at 93.6\% under \texttt{gpt-5.2-codex} is consistent with a ceiling, while SkillsBench still improves from 34.4\% to 36.8\%. Appendix~\ref{app:cost-accounting} reports the interaction and token accounting behind these runs.

\section{Conclusion}
\label{sec:conclusion}

We showed that the relational structure inside a skill library is worth more to an agent that can read and write it than to a retrieval pipeline that consumes it beforehand.

Two mechanisms carry that result independently. Letting the agent rank candidates itself, with no diffusion step over the graph, kept retrieval quality steady as the pool grew tenfold, where seeding-plus-diffusion did not. Letting it commit what execution taught it turned single observations into structure that transferred to held-out episodes and never evicted a skill the cold-start graph had already surfaced.

The practical implication is that a skill library should be built as writable structure behind an interface, not as a corpus to be indexed. Where the boundary falls between the retriever and the agent is a design choice rather than a fixed part of the stack, and moving it inside the agent returns a library that improves with use.

\section*{Limitations}

The propose-then-commit protocol is intentionally agent-driven: a single execution observation can suffice for the agent to commit an edge, bounded by structural guardrails (acyclicity, non-contradiction, reversibility) rather than a statistical confidence threshold (\S\ref{sec:online-edit}).
Aggregate edit effects are reported (\S\ref{sec:ablations}), but the long-horizon behaviour of single-observation self-evolution remains an open question we view as a direction for future work.

\clearpage
\bibliography{references}
\clearpage
\appendix
\section{Algorithms}
\label{app:algorithm}

Algorithm~\ref{alg:coldstart} builds the initial typed graph offline from the skill library (\S\ref{sec:cold-start}); Algorithm~\ref{alg:online} runs the per-episode online loop covering retrieval and graph edits (\S\ref{sec:graph-interface}, \S\ref{sec:online-edit}). The pair classifier $\psi$ returns one of \texttt{depends\_on}, \texttt{specializes}, \texttt{composes\_with}, \texttt{similar\_to}, or \texttt{none}; \texttt{conflicts\_with} is reserved for the online phase. Agent-facing calls (\texttt{search}, \texttt{show}, \texttt{propose-edge}, \texttt{edit-edge}) and the three commit-time invariants (acyclicity, non-contradiction, and append-only reversibility) appear inline in Algorithm~\ref{alg:online}.

\begin{algorithm}[H]
\caption{Cold-start graph construction.}
\label{alg:coldstart}
\small
\begin{algorithmic}[1]
\REQUIRE Skill library $\mathcal{C}$; embedder $\phi$; pair classifier $\psi$; per-ranking adaptive thresholds $\theta_{\text{self}}, \theta_{\text{needs}}$
\ENSURE Initial typed graph $G_0 = (V, E, \tau)$
\FOR{each $v \in \mathcal{C}$}
  \STATE $e_{\text{self}}(v) \gets \phi(\text{SKILL.md}(v))$
  \STATE $e_{\text{needs}}(v) \gets \phi(\text{HyDE-summary}(v))$ \COMMENT{$\bot$ if self-contained}
\ENDFOR
\STATE $P_{\text{self}} \gets \{(i,j) : \cos(e_{\text{self}}(i), e_{\text{self}}(j)) \ge \theta_{\text{self}}\}$
\STATE $P_{\text{needs}} \gets \{(i,j) : e_{\text{needs}}(i) \ne \bot \,\land\, \cos(e_{\text{needs}}(i), e_{\text{self}}(j)) \ge \theta_{\text{needs}}\}$
\STATE $E_0 \gets \emptyset$;\ \ $\tau_0 \gets$ empty map
\FOR{each $(a,b) \in P_{\text{self}} \cup P_{\text{needs}}$}
  \STATE $t \gets \psi(a, b)$
  \IF{$t \neq \texttt{none}$}
    \STATE $E_0 \gets E_0 \cup \{(a,b)\}$;\ \ $\tau_0(a,b) \gets t$
  \ENDIF
\ENDFOR
\STATE $G_0 \gets (\mathcal{C},\, E_0,\, \tau_0)$ \COMMENT{$\texttt{conflicts\_with}$ reserved for online phase}
\RETURN $G_0$
\end{algorithmic}
\end{algorithm}

\begin{algorithm}[H]
\caption{Per-episode online loop.}
\label{alg:online}
\small
\begin{algorithmic}[1]
\REQUIRE Initial graph $G_0$; task distribution $\mathcal{T}$; LLM agent $\pi$; budgets $K,D$
\ENSURE Evolved graph $G$ and append-only history log $H$
\STATE $G \gets G_0$;\ \ $H \gets \emptyset$
\FOR{each task $T \in \mathcal{T}$}
  \WHILE{episode not terminated}
    \STATE $\pi$ calls \texttt{search}/\texttt{show} as needed, then acts and observes outcome
    \IF{evidence supports an edit $\Delta$}
      \STATE $(\text{preview}, \text{history}) \gets \texttt{propose-edge}(G, \Delta)$
      \IF{\texttt{edit-edge} commits and $\textsc{Apply}(G,\Delta)$ passes invariants}
        \STATE $G \gets \textsc{Apply}(G, \Delta)$;\ \ append $\Delta$ to $H$ \COMMENT{reversible by recency}
      \ENDIF
    \ENDIF
  \ENDWHILE
\ENDFOR
\RETURN $G,\, H$
\end{algorithmic}
\end{algorithm}

\paragraph{Cold-start configuration.}
\begin{itemize}
\setlength{\itemsep}{0pt}\setlength{\parsep}{0pt}\setlength{\topsep}{2pt}
    \item Embedder: \texttt{text-embedding-3-}\allowbreak\texttt{large}
    \item LLM classifier: \texttt{gpt-5-nano} for pair classification
    \item Candidate pair threshold: adaptive (mean cosine similarity $+ 1\sigma$, clipped to $[0.35, 0.75]$)
    \item Maximum candidates per node: 12, with a minimum floor of 3
\end{itemize}

\section{Failure Analysis}
\label{app:failures}

\subsection{ALFWorld Failure Modes}

We categorise the 46 failures in the 140-episode MiniMax-M2.7 \skillDAG{} ALFWorld run of Table~\ref{tab:main-results}. Of these, 38 exhaust the 30-step cap.

\paragraph{Repeated no-op loops (32/46).} The agent cycles on actions the environment answers with ``Nothing happens'', most often trying to interact with an object that is not in the current room without recognising the need to navigate first. The agent cannot map the procedural instructions onto a valid environment action, and responds by issuing further \texttt{search} calls rather than by moving.

\paragraph{Step-budget exhaustion (8/46).} The agent explores correctly but runs out of steps: cleaning an object requires finding it, picking it up, navigating to a sink, using the sink, then navigating to the target receptacle.

\paragraph{Early termination (6/46).} The rest stop before the cap with few no-ops, in an incorrect final state; five of the six are state-change tasks where the agent placed the object without completing the state change.

\begin{table}[tb]
\centering
\footnotesize
\setlength{\tabcolsep}{4pt}
\begin{tabular}{@{}lr@{}}
\toprule
Task type & Success \\
\midrule
\texttt{look\_at\_obj\_in\_light}   & 13/13 (100\%) \\
\texttt{pick\_clean\_then\_place}   & 21/27 (78\%)  \\
\texttt{pick\_heat\_then\_place}    & 11/16 (69\%)  \\
\texttt{pick\_cool\_then\_place}    & 17/25 (68\%)  \\
\texttt{pick\_and\_place\_simple}   & 20/35 (57\%)  \\
\texttt{pick\_two\_obj\_and\_place} & 12/24 (50\%)  \\
\bottomrule
\end{tabular}
\caption{Success by ALFWorld task type over the same 140 episodes.}
\label{tab:failure-by-type}
\end{table}

Table~\ref{tab:failure-by-type} breaks the run down by task type. Single-step inspection never fails and two-object tasks fail half the time, but composition length is not a clean predictor on its own: \texttt{pick\_and\_place\_simple} is shorter than any state-change task and scores below all three.

\section{Edge Type Operational Definitions}
\label{app:edge-definitions}

We provide expanded operational definitions for each edge type, including the counterfactual tests that would ideally validate them:

\paragraph{\texttt{depends\_on}$(A, B)$.} Removing skill $B$ from the agent's available set drops the success rate of $A$ by a measurable margin. This captures prerequisite relationships: $B$ provides setup, context, or intermediate results that $A$ requires. Transitivity holds: if $A$ depends on $B$ and $B$ depends on $C$, then $A$ implicitly depends on $C$.

\paragraph{\texttt{composes\_with}$(A, B)$.} The joint success rate of tasks where both $A$ and $B$ are selected exceeds $\max(\text{success}(A), \text{success}(B))$ by a measurable margin. This captures synergistic combinations where skills complement each other. Symmetry holds. Transitivity does not, because composability is pairwise rather than chainable.

\paragraph{\texttt{similar\_to}$(A, B)$.} Skills $A$ and $B$ achieve comparable success rates on the same task distribution and can substitute for each other. Co-selecting both wastes context but does not cause harm. This is the weakest relationship type: it signals redundancy rather than interaction.

\paragraph{\texttt{conflicts\_with}$(A, B)$.} Co-selecting skills $A$ and $B$ causes a measurable drop in success rate compared to selecting either alone. This captures genuine interference: resource contention, contradictory instructions, state contamination, or protocol conflicts. Symmetry holds. Transitivity does not.

\paragraph{\texttt{specializes}$(D, A)$.} Skill $D$ achieves higher success than $A$ on a specific subdomain of $A$'s applicability, while inheriting $A$'s core functionality. When $D$ is available and the task falls in its subdomain, $D$ should be preferred over $A$. Transitivity holds within specialization hierarchies.

\section{Reproducibility Details}
\label{app:reproducibility}

\paragraph{Search budgets.}
All \texttt{search} calls use $K=5$ vector matches and BFS neighbor depth $D=2$. The agent may issue any number of \texttt{search} calls per episode and chooses query strings on its own schedule.

\paragraph{Models and embeddings.}
Agent runs use MiniMax-M2.7 (via the official MiniMax Anthropic-compatible endpoint), \texttt{gpt-5.2-codex} (via the OpenAI Codex CLI), and \texttt{deepseek-v4-flash}. Cold-start pair classification uses \texttt{gpt-5-nano}. Query embedding at runtime uses \texttt{text-embedding-3-large}, matching each \gos{} workspace's embedding configuration.

\paragraph{Step and attempt budgets.}
ALFWorld episodes are capped at 30 steps following \citet{li2026gos}. SkillsBench uses $n_{\text{attempts}}=1$ for the MiniMax cross-scale retrieval replay and $n_{\text{attempts}}=2$ for main task-reward results; the \texttt{gpt-5.2-codex} retrieval row replays the codex agent's queries under the same $n_{\text{attempts}}=2$ budget.

\paragraph{Online edit logging.}
Each accepted \texttt{edit-edge} writes an entry with \texttt{origin} (cold-start vs.\ online), \texttt{reason} (natural-language justification), and \texttt{task\_id} (the episode that produced the supporting evidence) into the graph's edge metadata.

\paragraph{Cold-start cost.}
At scale~1000 the cold-start construction takes approximately 200 \texttt{gpt-5-nano} calls in batched bucket mode (one call per anchor skill, classifying all candidate pairs in that bucket), plus one embedding call per skill for each of the two views ($e_{\text{self}}$ and $e_{\text{needs}}$).

\section{Interaction and Context Accounting}
\label{app:cost-accounting}

Table~\ref{tab:cost-accounting} compares the MiniMax \skillDAG{} and \gos{} ALFWorld runs on the same 140 tasks. Both use the same backbone, 30-step environment cap, evaluator, observation format, and action grammar; \skillDAG{} receives no larger environment-action allowance. The method-specific retrieval-command instructions differ.

\begin{table}[!t]
\centering
\footnotesize
\setlength{\tabcolsep}{3pt}
\begin{tabular}{@{}lrr@{}}
\toprule
Quantity & \skillDAG{} & \gos{} \\
\midrule
Environment actions   & 2{,}519 & 2{,}957 \\
Model-response turns  & 3{,}606 & 2{,}983 \\
Input tokens          & 21{,}054{,}686 & 8{,}650{,}556 \\
Output tokens         & 403{,}066 & 330{,}739 \\
\bottomrule
\end{tabular}
\caption{Observed MiniMax-M2.7 ALFWorld interaction and token accounting over 140 episodes. Input tokens include cache traffic.}
\label{tab:cost-accounting}
\end{table}

\skillDAG{} uses $2.43\times$ the cache-inclusive input tokens and $1.21\times$ the successful model-response turns while executing 85.2\% as many environment actions. Table~\ref{tab:retrieval-ops} breaks down the retrieval traffic behind those totals.

\begin{table}[!t]
\centering
\footnotesize
\setlength{\tabcolsep}{3pt}
\begin{tabular}{@{}lrr@{}}
\toprule
Quantity & \skillDAG{} & \gos{} \\
\midrule
Retrieval calls          & 273 (1.95) & 382 (2.73) \\
\quad automatic          & --         & 356 (2.54) \\
\quad agent-requested    & 273 (1.95) & 26 (0.19)  \\
\bottomrule
\end{tabular}
\caption{Retrieval operations over the same 140 MiniMax-M2.7 ALFWorld episodes. Totals with per-episode rates in parentheses. \skillDAG{} issues every retrieval itself; \gos{} retrieves automatically and additionally on request.}
\label{tab:retrieval-ops}
\end{table}

\section{Worked Example Subgraph}
\label{app:graph-example}

Edges are unweighted (present or absent); the LLM decides whether to add, remove, or retype an edge based on execution evidence, not statistical accumulation. Table~\ref{tab:edge-types} lists the runtime role of each type; the main-text Figure~\ref{fig:typed-edge-graph} shows a worked subgraph exercising all five.

\begin{table}[ht]
\centering
\begin{tabular}{@{}lp{4.4cm}@{}}
\toprule
\textbf{Type} & \textbf{Runtime role} \\
\midrule
\texttt{depends\_on} & Selecting $A$ pulls in $B$ as prerequisite \\
\texttt{composes\_with} & Surfaces composition opportunities \\
\texttt{similar\_to} & De-duplication: select one, not both \\
\texttt{conflicts\_with} & Prune: selecting $A$ excludes $B$ \\
\texttt{specializes} & Prefer specialized $D$ over general $A$ \\
\bottomrule
\end{tabular}
\caption{Edge type taxonomy; runtime roles.}
\label{tab:edge-types}
\end{table}

\section{Search Interface Field Semantics}
\label{app:interface}

\paragraph{Matches.} Cosine similarity is computed between the query embedding and each node's pre-computed embedding; the top-$K$ are returned, ranked by similarity.

\paragraph{Neighbors.} From the match set, BFS proceeds along walkable types (\texttt{specializes}, \texttt{composes\_with}, \texttt{depends\_on}, \texttt{similar\_to}) up to depth $D$, traversed bidirectionally so the agent sees both ends of a relationship (for \texttt{depends\_on}, both the prerequisite and any skill requiring the match as setup); the typed label travels with the edge. Each neighbor records its shortest distance from a match, predecessor, and edge type. \texttt{conflicts\_with} is not traversed. Skills already matched are excluded.

\paragraph{Conflicts.} All \texttt{conflicts\_with} edges incident to any match within one hop are returned separately. They have opposite polarity from neighbors: skills that should \emph{not} be co-selected. No transitive expansion (conflicts-of-conflicts are undefined).

\paragraph{Properties.} (i)~The agent issues queries on its own schedule and rephrases as understanding evolves, rather than receiving one episode-start bundle. (ii)~The three-field decomposition lets the agent drop a whole field under tight budget rather than truncating an opaque concatenation. (iii)~Edges carry an explicit type and reason, so the agent's trace documents selection/exclusion. After \texttt{search}, the agent issues \texttt{show} for chosen candidates; the decision is unconstrained (all matches, only typed neighbors, or skip the graph).

\section{Online-Edit Protocol Detail}
\label{app:online-edit}

The protocol separates inspection from mutation. \texttt{propose-edge} performs a dry-run returning what the change would do and surfacing all existing edges and recent history entries for the same skill pair, so the agent sees what prior episodes established before proceeding. \texttt{edit-edge} commits, appending a history entry with task ID, natural-language reason, and timestamp. The per-edit \texttt{reason} and \texttt{task\_id} log preserve auditability without gating the edit, placing graph-quality responsibility on the same agent that consumes it. The three commit-time invariants (acyclicity of the \texttt{depends\_on}/\texttt{specializes} backbone; non-contradiction, meaning no \texttt{conflicts\_with} edge on a pair that already carries a positive surfacing edge such as \texttt{depends\_on}, \texttt{specializes}, \texttt{composes\_with}, or \texttt{similar\_to}; and rollback by recency or task ID via the append-only log) are enforced on both \texttt{propose-edge} and \texttt{edit-edge}.

\section{Cold-Start Pipeline Detail}
\label{app:coldstart}

For each skill $v$, $e_{\text{self}}(v)$ embeds its identifier, description, and body preview; $e_{\text{needs}}(v)$ embeds an LLM prerequisite summary from two or three invoking tasks, omitting \texttt{self-contained} skills. Pairs exceeding $\theta=\max(0.35, \min(0.75, \mu + \sigma))$ ($\mu,\sigma$ are the mean/SD per ranking) under either $\cos(e_{\text{self}}(i), e_{\text{self}}(j))$ over all pairs or $\cos(e_{\text{needs}}(i), e_{\text{self}}(j))$ for non-empty needs are bucketed by lower-index anchor. One LLM call per bucket assigns a static type or \texttt{none}; accepted edges retain the label and reason.

\end{document}